\documentclass[lettersize,journal]{IEEEtran}
\usepackage{amsmath,amsfonts}
\usepackage{algorithmic}
\usepackage{algorithm}
\usepackage{array}
\usepackage[caption=false,font=normalsize,labelfont=sf,textfont=sf]{subfig}
\usepackage{textcomp}
\usepackage{stfloats}
\usepackage{url}
\usepackage{verbatim}
\usepackage{graphicx}
\usepackage{cite}
\usepackage{multirow}
\usepackage{bbding}
\usepackage{pifont}
\usepackage{wasysym}
\usepackage{amssymb}
 \usepackage{color}
 \usepackage{booktabs}
 \newcommand*{\method}{BOSS}
 \definecolor{myyellow}{rgb}{1,1, 0.6}
\definecolor{myorange}{rgb}{1, 0.8, 0.6}
\definecolor{myred}{rgb}{1, 0.6, 0.6}
\begin{document}

\title{BOSS: Bottom-up Cross-modal Semantic Composition with Hybrid Counterfactual Training  for Robust Content-based Image Retrieval}

\author{Wenqiao Zhang, Jiannan Guo, Mengze Li, Haochen Shi, Shengyu Zhang,\\ Juncheng Li,  Siliang Tang, Yueting Zhuang



\thanks{Wenqiao Zhang is with the College of Computer Science,
National University of Singapore, Singapore 117417 (e-mail: wenqiao@nus.edu.sg).}
\thanks{Haocheng Shi is with the College of Computer Science,
Universit\'{e} de Montr\'{e}al, Montr\'{e}al H3C 5J9 (e-mail: haochen.shi@umontreal.ca).}
\thanks{Jiannan Guo, Shengyu Zhang, Juncheng Li, Siliang Tang and Yueting Zhuang are with the College of Computer Science,
Zhejiang University, Hangzhou 310027, China (e-mail: jiannan@zju.edu.cn; sy$\_$zhang@zju.edu.cn; junchengli@zju.edu.cn; siliang@zju.edu.cn; yzhuang@zju.edu.cn).}
}

\markboth{Journal of \LaTeX\ Class Files,~Vol.~14, No.~8, August~2021}%
{Shell \MakeLowercase{\textit{et al.}}: A Sample Article Using IEEEtran.cls for IEEE Journals}


\maketitle
\begin{abstract}
Content-Based Image Retrieval (CIR) aims to search for a target image by concurrently comprehending the composition of an example image and a complementary text, which potentially impacts a wide variety of real-world applications, such as internet search and fashion retrieval. In this scenario, the input image serves as an intuitive context and background for the search, while the corresponding language expressly requests new traits on how specific characteristics of the query image should be modified in order to get the intended target image. This task is challenging since it necessitates learning and understanding the composite image-text representation by incorporating cross-granular semantic updates. In this paper, we tackle this task by a novel \underline{\textbf{B}}ottom-up cr\underline{\textbf{O}}ss-modal \underline{\textbf{S}}emantic compo\underline{\textbf{S}}ition (\textbf{BOSS}) with Hybrid Counterfactual Training framework, which sheds new light on the CIR task by studying it from two previously overlooked perspectives: \emph{implicitly bottom-up composition of visiolinguistic representation} and \emph{explicitly fine-grained correspondence of query-target  construction}. On the one hand, we leverage the implicit interaction and composition of cross-modal embeddings from the bottom local characteristics to the top global semantics, preserving and transforming the visual representation conditioned on language semantics in several continuous steps for effective target image search. On the other hand, we devise a hybrid counterfactual training strategy that can reduce the model's ambiguity for similar queries. This strategy resorts to the fine-grained correspondence construction of composed query and target image by providing extra explicit context-preserving and content-independent supervised signals. To verify the effectiveness of the proposed method, we conducted extensive experimental studies to show that our architecture BOSS outperforms existing techniques by achieving SOTA performance on two benchmark datasets.
\end{abstract}


\section{Introduction}

Along with the proliferation of visual data on the Internet, image search has become a critical component of a broad variety of downstream tasks that involves user experiences, such as product search~\cite{jing2008pagerank, guo2019attentive, liu2016deepfashion}, fashion retrieval~\cite{lin2020fashion,kuang2019fashion} and geolocalization~\cite{tian2020cross,lin2015learning}. Among many visual content search tasks, the most prevalent paradigm is taking an image or text as a query to search for a target image, commonly known as instance-level image retrieval and cross-modal retrieval. However, retrieving images solely based on a single uni-modal query~(\emph{e.g.}, text or image) might not be able to satisfy the fine-grained query intentions of users.
lajIn other words,  sophisticated information retrieval systems should enable users to express their intentions precisely from multiple facets, \emph{e.g.}, allowing users' interactive queries.

\begin{figure}[t]
\includegraphics[width=0.5\textwidth]{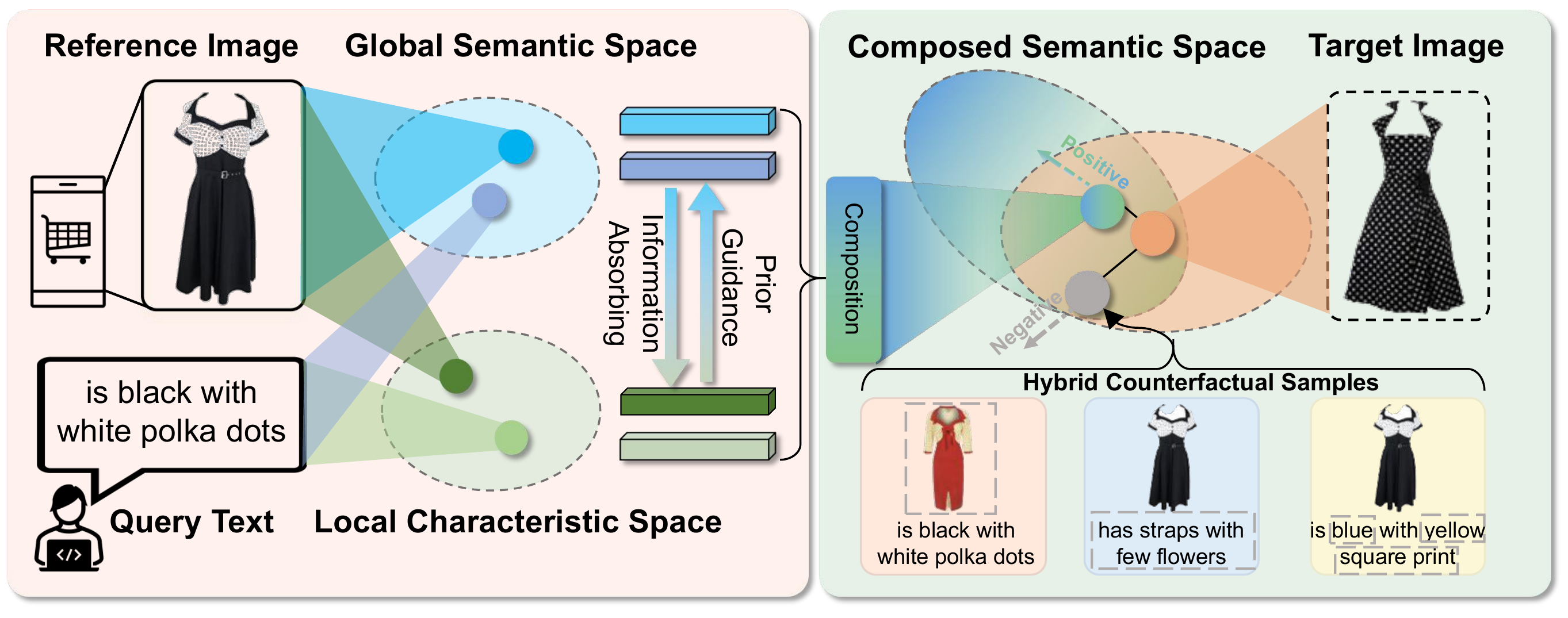}
\centering\caption{An example of content-based image retrieval and how our work facilitates this task. Green, blue, orange, and green-blue mixed dots represent the local-level, global-level, target image, and composition representation, respectively.  }
\label{fig1}
\vspace{-0.2cm}
\end{figure}

To alleviate the aforementioned restriction, a variety of approaches that refine the image retrieval process by incorporating users' interactive signals have been explored~\cite{han2017automatic,liu2016deepfashion, wu2004willhunter,mai2017spatial}.  Most of these methods formulate the retrieval refinement through \emph{text feedback} about spatial layouts~\cite{mai2017spatial} or relative attributes~\cite{han2017automatic,liu2016deepfashion, wu2004willhunter}. Most recently, TIRG~\cite{vo2019composing} introduced a novel task, namely \emph{Content-Based Image
Retrieval} (CIR): as shown in Figure~\ref{fig1}, a \emph{query text}  serves as an explicit request from the user, telling the retrieval system about \emph{what should change} and \emph{what should be preserved}  in \emph{reference image} to search the desired \emph{target image}. Such a multi-modal query setting provides a natural and flexible way for users to convey the detailed and precise specifications or modifications that they have in mind.


Broadly, most contemporary approaches~\cite{vo2019composing,hosseinzadeh2020composed, yang2021cross,chen2020image,kim2021dual,jandial2022sac,chen2020image} for CIR prefer to learn a  composed representation of the image-text query pair and then measure the similarities of the global representation with potential candidate images' features to search for the intended image. Despite intuitive, two indispensable characteristics for an ideal CIR model have been neglected by existing techniques:
(1) \textbf{Complementary feature utilization.} To obtain the composed representation of the multi-modal query, existing approaches mainly resort to cross-modality interaction and fusion operation on the global semantics (\emph{e.g.}, top layer features from encoder)~\cite{vo2019composing,yang2021cross,kim2021dual} and local characteristics (\emph{e.g.}, bottom layer features from encoder)~\cite{hosseinzadeh2020composed} or a simple separate/concatenate feature learning scheme ~\cite{jandial2022sac,wen2021comprehensive,chen2020image} of \emph{reference image} and \emph{query text} from their corresponding encoders (\emph{e.g.}, CNN or transformer). 
However, such a composing manner fails to capture the inherent relationships between informative characteristics from bottom layers and abstract-level semantics from top layers which with respect to the deep understanding perspectives: (a) 
\emph{detailed local characteristics} of the image region with the descriptive phrase (``white polka dots'') but without global semantic understanding of query. (b) \emph{abstract global semantics} (\emph{e.g.}, ``black and white dress'')  learned from increasing abstraction in a  \emph{hierarchical} order of encoders but lacks the concrete attributes rooted in different local positions. As a matter of fact, the two-level features act as the correlating and complementary understanding of multi-modal query, \emph{e.g.},  the local characteristics  can be regarded as the implicitly prior knowledge to guide the global semantic generation that learn the fine-grained image transformation and preservation. That is to say,  simply considering the aforementioned feature utilization without appropriately modeling the global-local composition of the  may lead to degenerate solutions. (2) \textbf{Linguistic snippet sensitivity.} One intrinsic challenge in CIR is to learn the transformation and  preservation of \emph{reference image} given the concretely requests among \emph{query text}.  To achieve this goal, a common learning way is that constructing the query-target correspondence for the desired image search, \emph{i.e.}, forcing the composed representation and \emph{target image} embedding closer.  However, such correspondence learning scheme suffers from the model's insensitivity of fine-grained linguistic variations in the \emph{query text}.  For example in Figure~\ref{fig1}, only replacing word  ``black'' $\rightarrow$ ``blue'' and   ``white'' $\rightarrow$ ``yellow'' in \emph{query text},  if the meanings of two queries are different, a robust CIR model should perceive the discrepancy and make different predictions.  Unfortunately, prior works based on simple query-target correspondence learning may be ambiguous or confused to identifying which image is the optimal matching target when input the attribute/object changing queries with similar semantic structure. In fact,   there are numerous contextually similar queries with different target image in CIR task, which may lead to coarse-grained and unsatisfactory search results. Summing up, these limitations push us to rethink the solution for the CIR.

Based on these insights,  we propose a novel framework for content-based image retrieval (CIR) task that called \underline{\textbf{B}}ottom-up cr\underline{\textbf{O}}ss-modal \underline{\textbf{S}}emantic compo\underline{\textbf{S}}ition (\textbf{BOSS}) with hybrid metric learning, which resolves the above issues in two major steps: (1) \textbf{Implicitly bottom-up semantic composition.} The key idea of this step is that, by leveraging the complementary global-local representation from different encoder's layers, the bottom-up synergistic composition of and visiolinguistic representation can be achieved,  which is crucial for the accurate CIR. Specifically, we first introduce a hierarchical representing strategy to enable the designed transformer based \emph{modification} and \emph{absorbing} blocks to capture multi-granular semantics across different levels (from local characteristics of the bottom layer to global semantics of the top layers). The visual contents preserving and transforming according to the given text are implicitly decomposed into three hierarchies (\emph{i.e.}, local characteristic composition, global-local absorbing composition and global semantic composition). As shown in Figure~\ref{fig1}, each step can be regarded as implicitly prior guidance to lead the subsequent composition inference with knowledge absorption. (2) ~\textbf{Explicitly fine-grained  correspondence learning.} 
We devise a hybrid counterfactual training (HCT) strategy that aims to encourage the CIR model to construct the fine-grained query-target correspondence for robust image retrieval. 
The HCST serves as a plug-and-play component to improve the CIR models’ query-sensitive abilities that consists of three different types of counterfactual samples (CSs) for a given query in Figure~\ref{fig1}: \textbf{ICS}, \textbf{TCS} and \textbf{CCS} are correspond to image-independent, text-independent and context-preserving CSs.
Such hybrid samples enable an explicitly bidirectional correspondence learning mechanism,  is helpful to establish the accurately one-to-one matching of composite query and desired image,  reducing the model's prediction uncertainty for similar queries.  Moreover, we incorporate an novel adaptive triplet loss, optimal transport-based domain alignment loss and reconstruct loss into the CIR learning, which can further incentive the accurate alignment of compositional query and target image. To summarise, our contribution is three-fold:


\begin{itemize}
\item  We tackle the challenging task of Content-Based Image Retrieval by learning the bottom-up cross-modal semantic composition via cascaded inference at the hierarchical level. This composition can capture the implicit visual modification and preservation by encapsulating the useful visiolinguistic information for effective image retrieval. 

\item A novel perspective is presented that enables the fine-grained image search by constructing the accurate query-target correspondence with a hybrid metric learning mechanism. 

\item The consistent superiority of the proposed BOSS is demonstrated on the two benchmark datasets that outperforms the existing SOTA methods. 
 \end{itemize}

\section{Related Work}

\noindent \textbf{Image Retrieval.}
In recent years, the development of deep learning brings prosperity to the field of computer vision~\cite{Li2022HERO,li2019walking,zhang2019frame,yang2021multiple,li2022compositional,li2020multi,zhang2022boostmis,li2022end,li2022dilated}. Among them,  Image Retrieval has drawn significant attention in this community.
Purely visual queries have been widely studied and are typically referred to as instance-level image retrieval (IIR)~\cite{philbin2007object}. With the rapid development of deep learning, IIR in the computer vision field~\cite{radenovic2016cnn,gordo2016deep,babenko2015aggregating,hou2018research,yue2015exploiting} has been actively studied. The prevailing IIR techniques often leverage global image representation~\cite{radenovic2016cnn,gordo2016deep,hou2018research}, facilitating query-target image matching, or extract local descriptors~\cite{babenko2015aggregating,tolias2016image,el2021training} for a costly yet robust matching. However, there still has been an intention gap between accurate user's intention and single image query for a long time. Therefore, cross-modal retrieval (CMR) is proposed to allow the users to express the search intention and search the desired image with the text or other modality queries. A broadly applied method is to map different modalities to a joint embedding space, and then use supervision information to guide the alignment of these modalities' embedding~\cite{kuang2019fashion,jing2020incomplete,gu2018look,wang2016comprehensive,wang2017adversarial,zhen2019deep}. However, the text is very abstract and sparse that is also unsatisfactory in conveying users' intention, making the application scenarios not extensive. 

\noindent \textbf{Interactive Image Retrieval.} Human-computer interactive methods prefer to use natural language in contemporary search engines. Thus, the Interactive Image Retrieval (IIR) is developed, a number of efforts (\emph{e.g.}, ~\cite{han2017automatic,liu2016deepfashion, wu2004willhunter,mai2017spatial}) have investigated effective ways to marry images with text representations. The text feedback can be provided in various ways, including absolute attributes (\emph{e.g.}, “blue”)~\cite{zhao2017memory,han2017automatic}, relative attributes (\emph{e.g.}, “more blue”)~\cite{han2017automatic,liu2016deepfashion, wu2004willhunter}, spatial layouts~\cite{mai2017spatial} and sketch~\cite{yu2016sketch,ghosh2019interactive}. Most recently, Content-Based Image Retrieval (\emph{CIR})~\cite{vo2019composing} is proposed. Following the setting that the input query is specified in the form of an image with a modification text (describes desired modifications to the reference image), many researchers devoted to capturing the joint expression of vision-language~\cite{jandial2022sac,anwaar2021compositional,vo2019composing,hosseinzadeh2020composed, yang2021cross,chen2020image,jandial2022sac}. Despite promising results, the above methods fail to model the interaction between local characteristics and global semantics, and suffer from the insensitive problem of similar contrastive querying. These constraints inspire us to develop the BOSS framework from two perspectives, \emph{i.e.}, implicitly bottom-up cross-modal semantic composition, and explicitly fine-grained correspondence enhancement.
\begin{figure*}[t]
\includegraphics[width=0.85\textwidth]{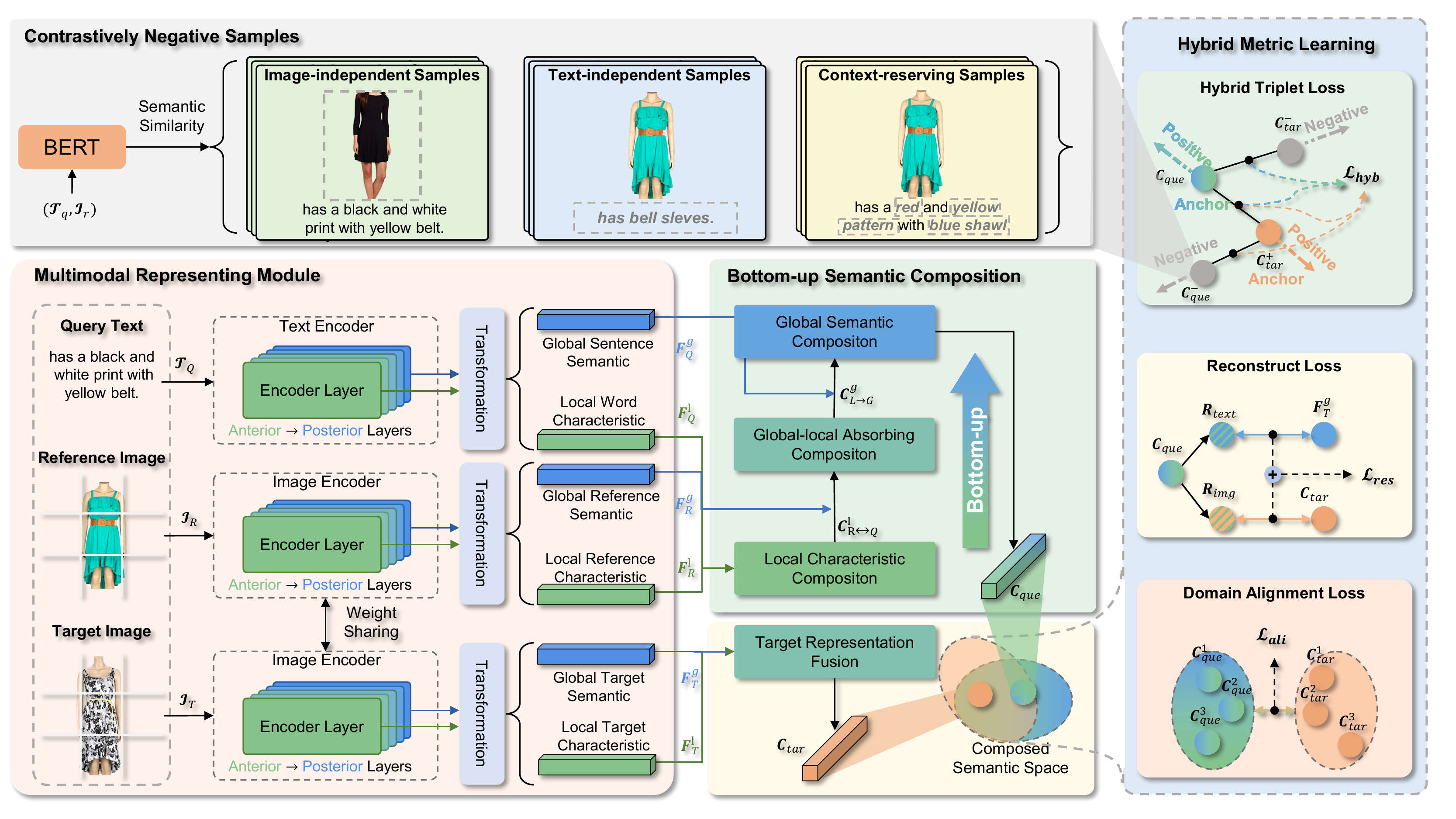}
\centering\caption{  $\textbf{Overview of  \method{}.}$ $\emph{Multimodal Representing Module}$  first encodes composite query and target image based on vision transformer and BERT at hierarchical levels. After this step, the $\emph{Bottom-up Semantic Composition}$  highlights salient regions in the image which need to be modified according to the text, by bottom-up inference at three levels. \emph{Hybrid Metric Learning} denotes the training objective, which can learn the fine-grained query-target correspondence for accurate CIR.}
\label{fig:framework}
\end{figure*}

\noindent \textbf{Compositional Representation Learning.} Many multimedia tasks, such as visual question answer (VQA)~\cite{antol2015vqa,lu2016hierarchical,li2019relation} and image captioning (IC)~\cite{huang2019attention,zhang2020relational,zhang2021consensus,cornia2020meshed,zhang2021tell}, is to learn a jointly compositional representation for image and text. These method often adopt  CNN~\cite{gu2018recent} and 
RNN~\cite{sundermeyer2012lstm} to encode the image and text, fusing the two embedding as a joint representation. With the development of transformers, the latest works leverage these advanced encoders to learn the latent multimodal representations~\cite{sur2020self,zhong2020self,yu2019multimodal,liu2021cptr}, mining richer semantic information. Notably, the modification text in the CIR fully expresses the editorial intent of the image. In other tasks, the image contents that do not conform to the text may be considered as background noise, while in our work these image contents may be the important preserved parts for reflecting the accurate users’ intention. Our method implicitly exploits the bottom-up compositional learning by hierarchical cross-modal semantic  Composition at three levels, modifying selectively the relevant image features and ensuring the preservation of the unaltered features.

 \section{Methodology}
This section describes the proposed Bottom-up crOss-modal Semantic compoSition  with Hybrid Metric Learning framework (BOSS) for content-based image retrieval (CIR). We will shed light on each module as well as the training strategy.

\subsection{Overview}
Before presenting our method, we first formulate the problem with some basic notions and terminologies. 
Given a \emph{reference image} $\mathcal{I}_R$ with the  \emph{query text} $\mathcal{T}_Q$, the goal of CIR is to identify the corresponding \emph{target image}  $\mathcal{I}_T$. This goal can be fulfilled by training a CIR model $\mathcal{M}$ to learn the matching pair (${(\mathcal{I}_R,\mathcal{T}_Q),\mathcal{I}_T}$). Here, $\mathcal{M}$ parameterized by $\Theta$  is trained to learn a image-text composite representation that is uniquely aligned with the visual representation of the target ground-truth image:
\begin{equation}
\begin{aligned}
\mathcal{M}({(\mathcal{I}_R,\mathcal{T}_Q),\mathcal{I}_T};\Theta)=\operatorname*{max}\limits_{\Theta}\kappa( \phi(\mathcal{I}_R,\mathcal{T}_Q),\psi (\mathcal{I}_T)))
\end{aligned}
\end{equation}
where  $\kappa(\cdot,\cdot)$ denote the similarity kernel, $\phi(\cdot)$ and $\psi(\cdot)$ are the composition encoder and the image encoder respectively.

Figure.~\ref{fig:framework} illustrates the overview of our  BOSS framework that contains two modules and one learning strategy: (a) \emph{Multimodal Representing Module} for vision
and language representation learning (in Sec.~\ref{sec:mrm});  (b) \emph{Bottom-up Semantic Composition} that absorbs local characteristic composition into global semantic composition learns at varying depths (in Sec.~\ref{sec:ccc}); (c)\emph{ Hybrid Counterfactual Training} presents the training strategy with multiple objectives for fine-grained image retrieval (in Sec.~\ref{sec:hml}). Besides, the pseudocode of the proposed BOSS framework for the CIR task is also provided in the appendix (Algorithm 1). We will further elaborate on our method for the CIR task as an example for clarity.

\subsection{Multimodal Representing Module}
~\label{sec:mrm}
This module aims to learn the representation of \emph{reference image} $\mathcal{I}_R$, \emph{target image} $\mathcal{I}_T$, and \emph{query text} $\mathcal{T}_Q$ at different levels for the subsequent bottom-up semantic composition.   

\subsubsection{ Multi-granular Visual Representation} 
We employ the Vision Transformer~\cite{dosovitskiy2020image} to derive a discriminative representation of the visual content of an image. Recent studies~\cite{tenney2019bert,vig2019multiscale} indicate that transformer encodes visual concepts with increasing abstraction, generally, becoming finer as we progress over levels. For the anterior layers, their outputs contain the basic syntactic information, which can be regarded as the local characteristic representation (\emph{LCR}). Thus, given the $\mathcal{I}_R$  and $\mathcal{I}_T$ $\in \mathbb{R}^{H \times W \times C}$,  we slice them into patches $\in \mathbb{R}^{K \times (P^2 \times C)}$, where  $K$ is the number of patches, $(H,W)$ is the input image resolution,  $(P,P)$is the resolution of each patch, and $C$ is the number of channels. Their  \emph{LCR} are extracted from the first layer of  transformer with  nonlinear projection, which denoted as $\textbf{F}_R^l$ $\in \mathbb{R}^{N \times d}$ and $\textbf{F}_T^l$ $\in \mathbb{R}^{N \times d}$,  respectively.

As the transformer inherently learns semantic concepts from the top orders, we extract the features of $\mathcal{I}_R$  and $\mathcal{I}_T$ from the posterior layer that can describe the more complex semantic meanings. Moreover, to further improve the representation quality, a learned linear projection maps the extracted features to the global semantic representation (\emph{GSR}),  $\textbf{F}_R^g$ $\in \mathbb{R}^{N \times d}$ and $\textbf{F}_T^g$ $\in \mathbb{R}^{N \times d}$,  respectively.

\subsubsection{Global-local Text Embedding}
The \emph{query text} $\mathcal{T}_Q$ is tokenized into a sequence of $M$ subword tokens by WordPiece~\cite{wu2016google}  as  in \verb+BERT+ ~\cite{devlin2018bert}.  We then prepend the 
special tokens \verb+[CLS]+ and \verb+[END]+ to the  sequence of textual subword tokens.
Similar to the image representing, we divide the \emph{query text} into two-level: local word embedding \emph{LWE}  and global sentence embedding \emph{GSE}. For \emph{LWE}, We assume the outputs of the first layer of \verb+BERT+ which sum each word embedding,  which can be calculated as the $\textbf{F}_Q^l$ $\in \mathbb{R}^{M \times d}$. 

Correspondingly, the contextual tokens have interacted with self-attention in multi-steps. The last layer of the \verb+BERT+ can represent the global-level meaning of the given $\mathcal{T}_Q$. Thus we concatenate these tokens with transformation to form the \emph{GSE} $\textbf{F}_Q^g$ $\in \mathbb{R}^{M \times d}$.
\subsection{Bottom-up Semantic Composition}
~\label{sec:ccc}
To learn the compositional representation that distinguishes \emph{what should change} and \emph{what should be preserved} in  $\mathcal{I}_R$ according to  $\mathcal{T}_Q$. We carefully design the Bottom-up Cross-modal Semantic Composition (BCSC) module, which can model the visiolinguistic representation at three-level cascaded inference. Before elaborating detailed process, we first introduce two composition methods:

\begin{itemize}
\item \textbf{Cross-modal Representation Modification} (CRM Block).  Given an embedding $\textbf{R}_{{R}}$ $\in \mathbb{R}^{N_k \times d}$ of \emph{Reference Modality R}  and a $\textbf{R}_{{Q}}$ $\in \mathbb{R}^{M_k \times d}$ of \emph{Query Modality Q}, this block seeks to learn the compositional embedding conditioned on  reference representation and query representation, obtained by selectively suppressing and highlighting the $\textbf{R}_{{R}}$ from query modality, is effective for the CIR learning.

\item \textbf{Representation Absorbing Composition} (RAC Block). The intuition of this module is to ``absorb'' the meaningful information from the local characteristic space instead of the simple ``feature fusion'', it can create an informative composition to robust the followup query-target matching. For instance, the local- and global-level representation,  denoted as $\textbf{R}_L$  $\in \mathbb{R}^{N_c \times d_c}$  and $\textbf{R}_G$ $\in \mathbb{R}^{N_c \times d_c}$, RAC  is expected to absorb the meaningful and discriminative information from $\textbf{R}_L$ , which served as prior guidance to $\textbf{R}_G$ for compositional representation generation.
 \end{itemize}
    
In this paper, the CRM and RAC blocks follow the recent advance in vision-language modeling technique, \emph{i.e}, transformer, to learn the visiolinguistic representation.  However, recent quantitative analyses~\cite{xu2021vitae} indicate the pure vision transformers are lacking convolutional inductive biases (e.g., translation equivariance~\cite{worrall2017harmonic}), which may lead the insensitive spatial locality in cross-modal modeling. To address this insue, we devise a novel pyramid pooling self-attention ($\mathcal{PSA}$) and pyramid pooling cross-attention ($\mathcal{PCA}$) mechanisms that redefine the multi-head self-attention ($\mathcal{MSA}$)~\cite{vaswani2017attention} to fully use the spatial locality of the visual semantics.  Given two representations $\textbf{R}_{A}$ $\in \mathbb{R}^{N_k \times d}$ and $\textbf{R}_{B}$ $\in \mathbb{R}^{M_k \times d}$  of modality A and B,  $\mathcal{PSA}$ and $\mathcal{PCA}$ are defined as below:

\begin{equation}
\begin{aligned}
&\mathcal{PSA}_{A}(\textbf{R}_{A})={\rm {softmax}}(\frac {\textbf{Q}_{A} \textbf{K}_{A}^{\top}} {\sqrt{d}_k}) \textbf{V}_{A}  
\\&=  {\rm softmax}(\frac {W_Q^A ( \textbf{R}_A + \sum_{i=2}^{N_p} {\rm Avg_i}(\textbf{R}_A)) \textbf{R}_A^{\top} {W_K^{A\top}}} {\sqrt{d}_k})\\& W_V^A\textbf{R}_A  \rightarrow  \mathcal{A}_{psa}  W_V^A \textbf{R}_{A} \rightarrow \hat{\textbf{R}}_{A},
 \end{aligned}
\end{equation}
\begin{equation}
\begin{aligned}
&\mathcal{PCA}_{B \rightarrow A}({\textbf{R}}_{ {B}}, {\textbf{R}}_{ {A}})={\rm softmax}(\frac {\textbf{Q}_B \textbf{K}_A^{\top}} {\sqrt{d}_k}) \textbf{V}_A \\& = {\rm softmax}
(\frac {W_Q^{BA} \textbf{R}_B (\textbf{R}_A + \ \sum_{i=2}^{N_p} {\rm Avg_i} (\textbf{R}_A))^{\top}  {W_K^{{BA}\top}}} {\sqrt{d}_k}) \\& W_V^{BA}\textbf{R}_A  \rightarrow  \mathcal{A}_{pca}  W_V^{BA} \textbf{R}_{A} \rightarrow \hat{\textbf{R}}_{BA},
\end{aligned}
\end{equation}
where $\{W_Q^A$, $W_K^A$, $W_V^A$, $W_Q^{BA}$, $W_K^{BA}$, $W_V^{BA}\}$ $\in \mathbb{R}^{d_k \times d_k}$ are  trainable weight matrices multiplied to $\textbf{R}_{A}$, $\textbf{R}_{A}$ is visual representation (\emph{e.g.}, embeddings of \emph{reference image} $\mathcal{I}_R$ or \emph{target image}  $\mathcal{I}_T$), $\textbf{R}_{b}$ denoted as the textual representation (\emph{e.g.}, embedding of \emph{query text} $\mathcal{T}_Q$). $\mathcal{A}_{psa}$ $\in \mathbb{R}^{N_k \times N_k}$ and $\mathcal{A}_{pca}$ $\in \mathbb{R}^{M_k \times N_k}$ are the attention matrices for computing the weighted average of  $\hat{\textbf{R}}_{A}$ and $\hat{\textbf{R}}_{BA}$ .
 $Avg_i$ makes average pooling for the $i \times i$ area centered with each patch of the given image. For edge patches, we only average the adjacent patches with the valid values, and ignore the padding patches.  Such novel blocks explicitly highlight the local correlation of the regions among one image, to effectively enhance the visually spatial sensitivity. 
 
With the two novel transform-based blocks, the CRM and RAC blocks carefully designed as follows:
\subsubsection{CRM Block}
\begin{figure*}[t]
\includegraphics[width=0.85\textwidth]{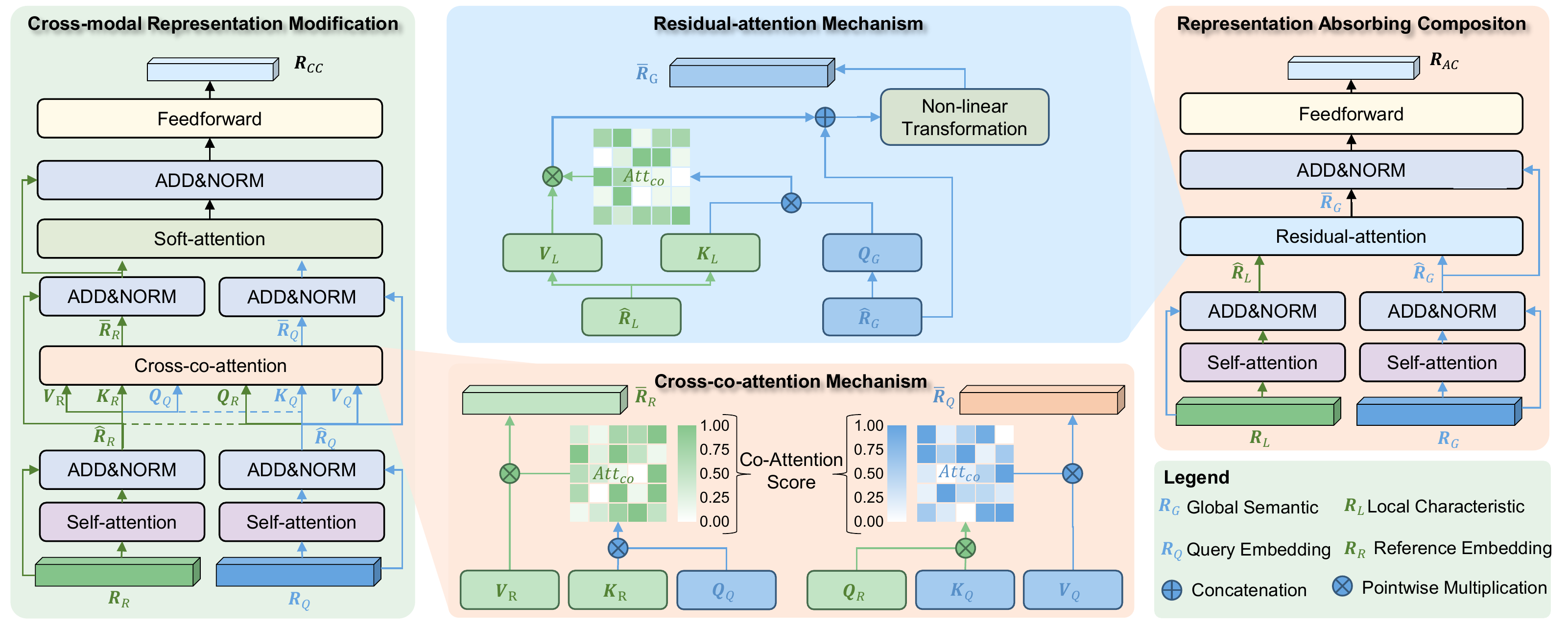}
\vspace{-0.2cm}
\centering\caption{Overview of cross-modal representation modification block and representation absorbing composition block.  Two blocks are responsible for \emph{transform} the
reference embedding and \emph{absorb} meaningful information from local characteristics. }
\label{fig:msa}
\end{figure*}

Figure~\ref{fig:msa} illustrated this block that consists of one self-attention layer, one bi-directional cross-attention layer and one soft-attention layer. To self-discover the latent region-to-region relationships essential for learning the transformation,  we feed them into the self-attention layer with layer normalization and residual connection: 
\begin{equation} \mathop{\hat{\textbf{R}}_M}_{M \in \{R, Q\}}\!\!\!\!=
\begin{cases} 
\mathcal{L}_n (\textbf{R}_{R}  \textcircled{+}  \mathcal{PSA}_{M}(\textbf{R}_{M}), & \mbox{If } \mathcal{J}(\textbf{R}_{M}) = \mathcal{V}\\
\mathcal{L}_n (\textbf{R}_{Q}  \textcircled{+}  \mathcal{MSA}_{M}(\textbf{R}_{M}), & \mbox{Others}
\label{computeA}
\end{cases}\!\!\!
\end{equation}
where $\mathcal{L}_n $ and \textcircled{+} are  layer normalization and residual connection, $\mathcal{J}(\cdot)$ is the function that judging the input representation is visual modality ($\mathcal{V}$). Thus, the self attended representation of reference and query modality $\hat{\textbf{R}}_{ {R}}$ and $\hat{\textbf{R}}_{ {Q}}$  are obtained after the self-spatial query.



Whilst self-attention captures the non-local correlations for feature transformation, the cross-modal correlations are still remains understudied, thus we introduce the  pyramid pooling cross-attention $\mathcal{CSA}$. $\overline{\textbf{{R}}}_{ {R}}$ and $\overline{\textbf{{R}}}_{ {Q}}$ can be obtained after the bidirectional cross-co-attention with  layer normalization and  residual connection.
\begin{equation} 
\begin{aligned}
\bar{\textbf{R}}_R=\mathcal{L}_n (\hat{\textbf{R}}_{ {R}} \textcircled{+} \mathcal{CSA}_{Q \rightarrow R} (\hat{\textbf{R}}_{ Q},\hat{\textbf{R}}_{ R}))\\
\bar{\textbf{R}}_Q=\mathcal{L}_n (\hat{\textbf{R}}_{ {Q}} \textcircled{+} \mathcal{CSA}_{R \rightarrow Q} (\hat{\textbf{R}}_{ R},\hat{\textbf{R}}_{ Q}))
\end{aligned}
\end{equation}

Next, to discover the latent relationships essentially of learned $\overline{\textbf{{R}}}_Q$ and $\overline{\textbf{{R}}}_R$ for image \emph{transformation} and  \emph{preservation}, we devote the soft-attention for $\overline{\textbf{{R}}}_R$  guided by $\overline{\textbf{{R}}}_Q$.  The final cross-modal composition is given by:

\begin{equation}
\begin{aligned}
\textbf{R}_{CC}= \mathcal{F}_w( \mathcal{L}_n(\overline{\textbf{{R}}}_R\textcircled{+}\mathcal{SOA}_{Q \rightarrow R}(\overline{\textbf{{R}}}_Q,\overline{\textbf{{R}}}_R))
)
\end{aligned}
\end{equation}
where $\textbf{R}_{CC}$ $\in \mathbb{R}^{N_k \times d_k}$,  $\mathcal{SOA}_{Q \rightarrow R}(.)$ is the soft attention mechanism.


\subsubsection{Representation Absorbing Composition}
The block includes one self-attention layer and one residual-attention layer, are built as a hierarchical order. The intuition of this module is that we want to ``absorb'' the meaningful information from the local characteristic space instead of the simple ``feature fusion'', it  can create an informative composition for follow-up CIR learning. For instance, the local- and global-level representation,  denoted as $\textbf{R}_L$  $\in \mathbb{R}^{N_c \times d_c}$  and $\textbf{R}_G$ $\in \mathbb{R}^{N_c \times d_c}$, the process of absorbing composition as follows:  
\begin{equation} \mathop{\hat{\textbf{R}}_M}_{M \in \{L, G\}}\!\!\!\!=
\begin{cases} 
\mathcal{L}_n (\textbf{R}_{M}  \textcircled{+}  \mathcal{PSA}_{M}(\textbf{R}_{M}), & \mbox{If } \mathcal{J}(\textbf{R}_{M}) = \mathcal{V}\\
\mathcal{L}_n (\textbf{R}_{M}  \textcircled{+}  \mathcal{MSA}_{M}(\textbf{R}_{M}), & \mbox{Others}
\label{computeA}
\end{cases}\!\!\!
\end{equation}


After the self-attention modeling, the intermediate embeddings are  generated by $\hat{\textbf{R}}_L$ and $\hat{\textbf{R}}_G$.  In this setting, we assume that the global semantic representation has higher weight, and $\hat{\textbf{R}}_L$ can be regarded as the ``prior guidance'' for composition generation. Thus,
we leverage the residual-attention mechanism that queries the meaningful information from $\hat{\textbf{R}}_L$ guided by $\hat{\textbf{R}}_G$ as below:
\begin{equation}
\begin{aligned}
\!\!\!\!\mathcal{REA}_{G \rightarrow L}(\hat{\textbf{R}}_G, \hat{\textbf{R}}_L)= 
\mathcal{T}_{l}[\mathcal{CSA}_{G \rightarrow L}(\hat{\textbf{R}}_G, \hat{\textbf{R}}_L),\hat{\textbf{R}}_G] \rightarrow \overline{\textbf{R}}_G\!\!
\end{aligned}
\end{equation}
where $[\cdot,\cdot]$  is the concatenate operation,  $\mathcal{T}_{l}$  is the non-linear transformation. In  this formula, the attentive representation through cross-co-attention is absorbed by the non-linear projection as $\overline{\textbf{R}}_G$.  

Lastly,  the fused embedding $\overline{\textbf{R}}_G$ performs the residual connection and layer normalization, then input to the feedforward layer to produce the final absorbing composition:
\begin{equation}
\begin{aligned}
\textbf{R}_{AC}= \mathcal{F}_w( \mathcal{L}_n(\hat{\textbf{R}}_G\textcircled{+}\overline{\textbf{R}}_G))
\end{aligned}
\end{equation}
where $\textbf{R}_{AC}$ $\in \mathbb{R}^{N_c \times d_c}$. This composition representation absorbs useful knowledge from local characteristics, which can enhance the query-target matching accuracy.  
\subsubsection{ Hierarchical  Composition Inference}
To gradually digest the information flows from vision and language domains,  we propose the bottom-up composition that learn the visiolinguistic representation at continuous three major sub-parts: 
\begin{itemize}
\item \textbf{Local Characteristic Composition}.  There are \emph{transformation} and \emph{preservation} in lots of paired image-text cases have local characteristic correlations in CIR. To  handle the ``how to change''  in the \emph{reference image}, we utilize the Cross-modal Representation Modification (\emph{CRM}) block to learn the local characteristic composition $\textbf{C}^l_{R\leftrightarrow Q}$ on the basis of  local-level   $\textbf{F}_R^l$  and  $\textbf{F}_Q^l$ at the first step.

\item \textbf{Global-local Absorbing  Composition}. This step is expected to absorb the meaningful and discriminative information from $\textbf{C}^l_{R\leftrightarrow Q}$, which served as prior guidance from the local characteristic level to robust the follow-up composition-target matching. Therefore, the latent embedding of global-local absorbing composition $\textbf{C}^{g}_{L \rightarrow G}$ is reasoned by the Representation Absorbing Composition (\emph{RAC}) module.

\item \textbf{Global Semantic  Composition}. Eventually, in order to model the final composition $\textbf{C}^{g}_{L \rightarrow G}$ from vision and language domains, the output of this stream is updated by aggregating the essential semantics from the intermediate representation $\textbf{C}^{g}_{L \rightarrow G}$ and \emph{query text} global semantic latent vector $\textbf{F}_Q^g$.
 \end{itemize}


\begin{equation}
\begin{aligned}
&\underbrace{CCM_{R\leftrightarrow Q}^l (\textbf{F}_R^l, \textbf{F}_Q^l)}_{{\rm 1st \ \ Composition}} \rightarrow \underbrace{RAC_{L \rightarrow H}^q (\textbf{C}^l_{R\leftrightarrow Q}, \textbf{F}_R^g)}_{{\rm 2nd \ \ Composition}}
 \\&
\rightarrow  \underbrace{CCM_{R\leftrightarrow Q}^h (\textbf{C}^{g}_{L \rightarrow G}, \textbf{F}_Q^h)}_{ {\rm 3rd \ \ Composition}} \rightarrow \textbf{C}^{g}_{L \rightarrow G}
\end{aligned}
\end{equation}
where $\textbf{C}^{g}_{L \rightarrow G}$ $\in \mathbb{R}^{N \times d}$.  In summary, the final composite embedding $\textbf{C}_{que}$, reasoned from such bottom-up composition at hierarchical levels, is effective in capturing the implicit visual \emph{modification} and \emph{preservation} in \emph{reference image} depending on the text modifier.

\noindent$\textbf{Target Representation Fusion.}$ 
We also employ the \emph{CCM}  to fuse the local characteristic $\textbf{F}_T^l$ and global semantic $\textbf{F}_T^h$ of \emph{target image}:
\begin{equation}
\begin{aligned}
\textbf{C}_{tar}=RAC_{L \rightarrow G}^t (\textbf{F}_T^l, \textbf{F}_T^g)
\end{aligned}
\end{equation}
where $\textbf{C}_{tar}$ $\in \mathbb{R}^{N \times d}$. Such design broadcasts the local information from $\textbf{F}_T^l$ required to the latent embedding $\textbf{F}_T^g$ that characterize the jointly target composite representation $\textbf{C}_{tar}$. 
\subsection{Hybrid Counterfactual Training}
~\label{sec:hml}
In CIR task, a common adopted learning way, deep metric learning (\emph{DML})  that aims to construct the query-target correspondence for the desired image search, \emph{i.e.}, forcing the \emph{reference image} embedding and \emph{target image} embedding closer, while pulling apart the representations of mismatched query-image pairs. However, such \emph{na{\"i}ve}  one-to-one paired learning scheme cannot guarantee the target representation is far away from the negative composed query, especially the contextually similar queries with  linguistic variations, which may hurt the model's sensitivity for negative composition. To address this problem, we introduce a plug-and-play component called Hybrid Counterfactual Training (\emph{HCT}) strategy to model the fine-grained query-target correspondence based on  three different types of counterfactual samples:

\begin{itemize}
\item \textbf{Image-independent and Text-independent Counterfactual Sample} (\emph{ICS} and \emph{TCS}). On the one hand, to enhance the model's sensitivity for the irrelevant visual and textual samples, first, we aims to seek the ICS and TCS. To enhance the model's sensitivity for the irrelevant visual query. Specifically, give the \emph{reference image} $\mathcal{I}_R$ with corresponding \emph{query text} $\mathcal{T}_Q$, we leverage \verb+BERT+ as a pretrained bidirectional language model to find the least relevant texts $\{\mathcal{T}^{i}_{N_q}|_{i=1}^{K_q}\}$  and its corresponding images $\{\mathcal{I}^{i}_{N_r}|_{i=1}^{K_r}\}$  according to semantic similarity.  Thus, those texts and images combine the original \emph{reference image} and \emph{query text} to construct the \emph{ICS} $(\{\mathcal{I}_R^i|_{i=1}^{K_q}\}, \{\mathcal{T}^{i}_{N_q}|_{i=1}^{K_q}\})$ and \emph{TCS} $(\{\mathcal{I}^{i}_{N_r}|_{i=1}^{K_r}\}, \{\mathcal{T}_q^i|_{i=1}^{K_r}\})$.

\item \textbf{Context-preserving Counterfactual Sample} (\emph{PCS}). On the other hand, the CIR model lacks the fine-grained query-target correspondence due to it is especially insensitive for the semantically similar queries, \emph{i.e.}, the context-preserving samples (attribute-related negative \emph{query text}). To relieve this issue, we extract these queries as negative samples to improve the sensitivity. However, it is difficult to search the context-preserving queries due to it requires semantic understanding of each word in a reference text. In our approach, such negative sample extracting consists of two steps: First,  we generated the preliminary candidates by masking the attribute words (\emph{e.g.}, adjective and noun) of \emph{query text} $\mathcal{T}_Q$ according to the property provided by NLTK~\cite{loper2002nltk} and replace them with random words (same word property) to obtain $K_1$ samples. Second, we pass the original $\mathcal{T}_Q$ into \verb+BERT+ to compute semantic similarity, and finally, choose the top $K_2$ texts with the \emph{reference image} as the context-preserving queries.
\begin{equation}
\begin{aligned}
(\{\mathcal{I}_R^i|_{i=1}^{K_2}\}, \{\mathcal{T}^{i}_{N_c} |_{i=1}^{K_2}\})=\mathcal{T}_{K_2}(\mathcal{P}_s(\{\mathcal{T}^{i}_{N_c} |_{i=1}^{K_1}\} | \text{BERT}(\mathcal{T}_Q))
\end{aligned}
\end{equation}
where $(\{\mathcal{I}_R^i|_{i=1}^{K_2}\}, \{\mathcal{T}^{i}_{N_c} |_{i=1}^{K_2}\})$ are selected context-preserving negative samples,  $\mathcal{P}_s$ measures the probabilities predicted by \text{BERT},  $\mathcal{T}_{K_2}$ indicate that the selected number at second stage.
 \end{itemize}


\subsubsection{Bidirectional Triplet Loss}
As mentioned earlier, our goal is to construct the fine-grained query-target correspondence of input query and \emph{target image} based on these contrastively negative samples that extracted at two degrees.
Thus, we introduce the bidirectional triplet loss as our primary objective to ensure the semantic matching of composition and the target representation with high similarity. The hybrid loss $\mathcal{L}_{hyb}$ and triplet loss $\mathcal{L}_{tri}$ are defined as follows:
 \begin{equation}
 \begin{aligned}
&\!\!\!\mathcal{L}_{tri}(\textbf{X}, \textbf{Y}, m)= max(0, \rvert\rvert \textbf{X}^+ -\textbf{Y}\rvert\rvert_2-\rvert\rvert\textbf{X}^- -\textbf{Y}\rvert\rvert_2 +m)\!\!\!
\\
&\mathcal{L}_{bid}(\textbf{C}_{que}, \textbf{C}_{tar}, m, m_a)=\lambda_q \mathcal{L}_{tri} (\textbf{C}_{que}, \textbf{C}_{tar},m) \\&+\lambda_t \mathcal{L}_{tri}(\textbf{C}_{tar}, \textbf{C}_{que}, m_a)\\
\end{aligned}
\end{equation}
where $\textbf{X}^+$ and $\textbf{X}^-$ denotes the positive and negative samples, $\lambda_q$ and $\lambda_t$ are the weight hyper-parameters $\rvert\rvert \cdot \rvert\rvert_2$ is the \emph{L2} distance.
 \begin{equation}
 \begin{aligned}
m_a=\frac{1-a^{s_{qt}}}{1-a}m
\end{aligned}
\end{equation}
where  $s_{tq}$ indicates the semantic similarity of $\textbf{C}_{que}$, $\textbf{C}_{tar}$,  $a$ is hyperparameter to control the adaptive margin $m_a$. If $s_{qt}$ is close to 0,  $m_a$ will be assigned a large value and a small value
otherwise. Thus, the adaptive margin for counterfactual training can be achieved.

\subsubsection{Reconstruct Loss} 
This loss constraints visual and linguistic projections of $\textbf{C}_{que}$, denoted by $\textbf{R}_{img}$ and $\textbf{R}_{text}$, to align with latent embeddings $\textbf{C}_{tar}$ and $\textbf{F}_T^g$, respectively. This objective by reconstruction regularizes and reinforces the balanced utilization of both text and image in composed embedding $\textbf{C}_{que}$.
 \begin{equation}
 \begin{aligned}
\mathcal{L}_{res}=\lambda_{img} \rvert\rvert \textbf{R}_{img}-\textbf{C}_{tar} \rvert\rvert_2+ \lambda_{text} \rvert\rvert \textbf{R}_{text}-\textbf{F}_T^g \rvert\rvert_2
\end{aligned}
\end{equation}
where  $\lambda_{img}$ and $\lambda_{text}$ are pre-defined hyperparameters.
\subsubsection{Domain Alignment Loss} 
To further learn the fine-grained semantic correspondence of the composition domain and \emph{target image} domain, we propose to use optimal transport (\emph{OT}) ~\cite{peyre2019computational} for aligning representation distributions of different domains to bridge their gap. The training process with \emph{OT} consists of two steps. First, the \emph{OT} calculates the cost matrix $\mathcal{C}_m$ between the feature distributions of the composition domain and target domain. Second, \emph{OT} plan assigns each feature to the features from the other modality with different weights, which can be viewed as a prediction for correspondence relationship. In detail, the  Wasserstein Distance~\cite{shen2018wasserstein} (\emph{WD}) is employed to align the composition domain to the target domain, the \emph{WD} $W(X, Y)$ and alignment loss can be defined as:
 \begin{equation}
\begin{aligned}
&\mathcal{W}_d(\textbf{C}, \textbf{T}) = \inf_{\gamma \in \Pi (\textbf{C}, \textbf{T})} \mathbb{E}_{\textbf{c} \sim\gamma, \textbf{t} \sim\gamma } <(\textbf{c},\textbf{t}),\mathcal{C}_m(\textbf{c},\textbf{t})>_F\\
&\mathcal{L}_{ali}=\lambda_a \mathcal{W}_d(\textbf{C}, \textbf{T}) 
\end{aligned}
\end{equation}
where  $<\cdot,\cdot>_F$ is the Frobenius dot product, \textbf{C} and \textbf{T} are all the embeddings for the composition domain and target image domain.

\begin{table*}
\caption{Quantitative results of content-based image retrieval on FashionIQ dataset. Superscript $^{\dagger}$ denotes the method variant with different \texttt{backbone}.  Avg: averaged RC@10/50 computed over three categories. Overall \textcolor{red}{$1^{st}$} and \textcolor{blue}{$2^{nd}$} best in \textcolor{red}{red}/\textcolor{blue}{blue}.}
\centering
\begin{tabular}[width=1\textwidth]{l|c|cc|cc|cc|cc|c}
\toprule[1.5pt]
\multicolumn{1}{c|}{\multirow{2}{*}{\textbf{Methods}}}&\multicolumn{1}{c|}{\multirow{2}{*}{\textbf{BERT}}}  &\multicolumn{2}{c|}{\textbf{Dress} } &\multicolumn{2}{c|}{\textbf{Toptee}}   & \multicolumn{2}{c|}{\textbf{Shirt} } &\multicolumn{2}{|c}{\textbf{AVG} } & \multicolumn{1}{|c}{\multirow{2}{*}{\textbf{Total AVG}}}\\            
  \cmidrule(l){3-4} \cmidrule(l){5-6} \cmidrule(l){7-8} \cmidrule(l){9-10} 
\multicolumn{1}{c|}{} & &\multicolumn{1}{c}{\textbf{RC@10}} & \multicolumn{1}{c|}{\textbf{RC@50}}  & \textbf{RC@10} & \textbf{RC@50}     & \textbf{RC@10} & \textbf{RC@50}        & \textbf{RC@10} & \textbf{RC@50} \\ 
 \midrule[0.8pt]
  \midrule[0.8pt]
  \textbf{FILM}~\cite{perez2018film} &&14.23& 33.34& 17.30& 37.68& 15.04 &34.09& 15.52& 35.04 &25.28\\
\textbf{TRIG}~\cite{vo2019composing}&&14.87 &34.66& 19.08& 39.62& 18.26& 37.89&27.40 &37.39 &32.40\\
\textbf{Relationship}~\cite{santoro2017simple}&&15.44 &38.08 &21.10 &44.77 &18.33 &38.63 &18.29 &40.49&29.39\\
\textbf{VAL}~\cite{chen2020image} &&21.47& 43.83& 26.71& 51.81 &21.03& 42.75& 23.07 &46.13&34.60 \\
\textbf{RTIC}~\cite{shin2021rtic} &&
\textcolor{blue}{\textbf{28.21}} &51.41 &28.00 &55.58 &21.30 &44.80 & 25.83& 50.59&38.22\\
\textbf{ARTEMIS}~\cite{delmas2022artemis}  &&24.84&23.63& 20.40& 49.00&47.39&43.22& 22.95&46.54&34.75
\\
\textbf{ARTEMIS}$^{\dagger}$~\cite{delmas2022artemis} &&27.16& \textcolor{blue}{\textbf{52.40}} & 29.20& 54.83 &21.78 &43.64& 26.05& 50.29&40.44 
\\ 
\textbf{SAC}~\cite{jandial2022sac}&&
26.13& 52.10 &31.16& 59.05& 26.20 &50.93& 27.83& 54.03& 40.93\\

 \midrule[0.8pt]
\textbf{ComposeAE}~\cite{anwaar2021compositional} &\CheckmarkBold& 14.03& 35.1& 15.8 &39.26& 13.88 &34.59& 19.89& 36.31&25.44\\
\textbf{SAC}$^{\dagger}$~\cite{jandial2022sac} &\CheckmarkBold &26.52& 51.01& \textcolor{blue}{\textbf{32.70}}& \textcolor{red}{\textbf{61.23}}& \textcolor{blue}{\textbf{28.02}}& \textcolor{blue}{\textbf{51.86}}& \textcolor{blue}{\textbf{29.08}}& \textcolor{blue}{\textbf{54.70}}&\textcolor{blue}{\textbf{41.89}}\\
 \midrule[0.8pt]
\textbf{BOSS} (ours) &\CheckmarkBold&\textcolor{red}{\textbf{29.05}}&\textcolor{red}{\textbf{54.41}}&\textcolor{red}{\textbf{34.73}}&\textcolor{blue}{\textbf{60.08}}&\textcolor{red}{\textbf{31.04}}&\textcolor{red}{\textbf{53.78}}&\textcolor{red}{\textbf{31.61}}&\textcolor{red}{\textbf{56.09}}&\textcolor{red}{\textbf{43.85}}\\
\bottomrule[1.5pt]
\end{tabular}
\label{tab:results_1}
\end{table*}

\begin{table}
\caption{\textbf{Quantitative comparison on  Fashion200k dataset.} }
\centering
\begin{tabular}[width=1\textwidth]{l|ccc|c}
\toprule[1.5pt]
\multicolumn{1}{c|}{\textbf{Methods}} & \multicolumn{1}{c|}{\textbf{RC@1}} & \multicolumn{1}{c|}{\textbf{RC@10}}  & \textbf{RC@50} & \textbf{AVG}
\\  \midrule[0.8pt]
  \midrule[0.8pt]
\textbf{FILM}~\cite{perez2018film}&12.9 &39.5 &61.9&38.1  \\
\textbf{TRIG}~\cite{vo2019composing}&14.1 &42.5 &63.8&40.1\\
\textbf{Relationship}~\cite{santoro2017simple}&13.0 &40.5 &62.4&43.0 \\
\textbf{VAL}~\cite{chen2020image} &\textcolor{blue}{\textbf{22.9}} &\textcolor{blue}{\textbf{50.8}} &\textcolor{blue}{\textbf{72.7}}&\textcolor{blue}{\textbf{48.8}} \\

\textbf{ARTEMIS}$^{\dagger}$~\cite{delmas2022artemis}& 21.5& 51.1& 70.5& 47.7\\
\textbf{ARTEMIS}~\cite{delmas2022artemis}&20.2& 49.3& 69.3& 46.2 \\

\midrule[0.8pt]
\textbf{ComposeAE}~\cite{anwaar2021compositional} & 16.5& 45.4& 63.1&41.7\\
\midrule[0.8pt]
\textbf{BOSS} (ours) &\textcolor{red}{\textbf{20.1}} &\textcolor{red}{\textbf{59.6}} &\textcolor{red}{\textbf{79.5}}& \textcolor{red}{\textbf{53.1}}\\
\bottomrule[1.5pt]
\end{tabular}
\label{tab:results_2}
\end{table}


\section{Experiments}
We benchmark \method{} for CIR on  three datasets to verify its effectiveness, and then discuss \method{} ’s property with controlled studies.
\subsection{Dataset and Setting}
\noindent \textbf{Datasets.} \textbf{FashionIQ}~\cite{wu2021fashion} is composed of 46.6k training images and around 15.5k images for both the validation and test sets, covering three categories: \texttt{Dresses}, \texttt{Top\&tees} and \texttt{Shirts}. The dataset is characterized by medium support text descriptions with an average length of 10.69 words per sample. \textbf{Fashion200k}~\cite{han2017automatic} is a large-scale fashion dataset crawled from multiple online shopping websites, which consists of 200k images of five different fashion categories, namely: \texttt{pants}, \texttt{skirts}, \texttt{dresses}, \texttt{tops} and \texttt{jackets}. We incorporate our BOSS and use the training split of around 172k images for training and the testset of 33k test queries for evaluation. \textbf{Shoes}~\cite{berg2010automatic} is a dataset originally crawled from \emph{like.com}. This dataset includes 14,658 images of footwear tagged with relative captions for dialog-based interactive retrieval.Following~\cite{guo2018dialog}, we use 10,000 training samples for training and 4,658 test samples for evaluation.



\noindent \textbf{Comparison of Methods.}  We compare the results of \method{} with a wide range of baselines including early works and recent State of the Art models on this task: File~\cite{perez2018film}, \textbf{TRIG}~\cite{vo2019composing}, \textbf{Relationship}~\cite{santoro2017simple}, \textbf{VAL}~\cite{chen2020image}, \textbf{RTIC}~\cite{shin2021rtic},  \textbf{ComposeAE}~\cite{anwaar2021compositional}, \textbf{SAC}~\cite{jandial2022sac} and \textbf{ARTEMIS}~\cite{delmas2022artemis}. More details of baselines are explained in the appendix.

\noindent \textbf{Implementation\&Evaluation Details.} We employ the pretrained Vision Transformer~\cite{dosovitskiy2020image} and \texttt{BERT}~\cite{devlin2018bert} as the backbone for the initial image and text feature extractor.  Feature dimension $\textbf{d}$ in our experiments is project to 768. We measure the retrieval performance with common metrics in information retrieval, including Recall at K(RC@K and K=1, 10, 50). RC@K
is the percentage of test queries that at least one relevant item is found among the top-K retrieved results. The initial learning rate is set to 2e-5 and the network is optimized by AdamW~\cite{loshchilov2017decoupled} optimizer. The
10\% proportion of warm up and cosine decay are used for scheduling the learning rate.  Besides, we use an identical set of hyperparameters ($B$=32, $ME$=100, $\lambda_q$=1, $\lambda_t$=0.4, $\lambda_{img}$=0.1, $\lambda_{text}$=1 and $\lambda_{a}$=0.01) \footnote{$B$ and $ME$ refers to the batch size and max training epoch. }.

\subsection{Experimental Results}
We now compare BOSS to SOTA approaches on each benchmark dataset and then discuss the experimental results.

\noindent \textbf{Quantitative Results.} Table~\ref{tab:results_1}, ~\ref{tab:results_2} and ~\ref{tab:results_3} summarize the quantitative results of our framework and baselines on FashionIQ, Fashion 200K and Shoes dataset.  From the these tables, we have the following findings: (1) In general, BOSS achieves the best performance on almost all the metrics to SOTA on FashionIQ and Fashion200K dataset. In particular, \method{} outperforms other baselines in terms of RC@10 and RC@50 by a large margin (FashionIQ: \textbf{\underline{2.53\% $\sim$ 16.09\%}} and \textbf{\underline{1.39\% $\sim$ 21.05\%}}; Fashion200K: \textbf{\underline{6.8\% $\sim$ 20.1\%}} and \textbf{\underline{6.8\% $\sim$ 17.6\%}}  ) for CIR task.
These results demonstrate the superiority and generalizability of our proposed model. (2) It is worth noting that the performance growth of RC@10 results goes beyond the RC@50 when compared with baselines on all the datasets. This phenomenon is reasonable as the Hybrid Counterfactual Training (\emph{HCT}) improves the model search sensitivity for fine-grained composition and \emph{target image} matching. Therefore, BOSS can reflect its superiority in selecting the accurate image with fewer candidate recommendations. Similar results appear in the Shoes dataset, we achieve the \underline{\textbf{highest score on RC@1 metric}}. The results once again prove that our model can build the fine-grained query-target correspondence that accurately locates the desired image. We conduct in-depth analysis of \emph{HML} in section~\ref{sec:ida}. (3) From table~\ref{tab:results_2}, we observe that BOSS has not outperformed SOTA but achieves comparable results on the Shoes dataset.
According to further analysis, the possible reason is that the data scale of Shoes dataset (14,658)  is relatively smaller compared with other datasets.
As commonly known,  transformers lack some of the \emph{inductive biases} inherent to CNNs, such as translation equivariance and locality, and therefore do not generalize well when trained on insufficient amounts of data. The training data of Shoes may not be sufficient to train a complex transformer-based deep model. On the contrary, while the Fashion200K dataset (200k images) is about 14 times larger than the Shoes dataset, our method far surpasses the existing methods on all metrics, which shows the BOSS's effectiveness on larger datasets.

\begin{table}
\caption{\textbf{ Quantitative comparison  on Shoes dataset.} }
\centering
\begin{tabular}[width=1\textwidth]{l|ccc|c}
\toprule[1.5pt]
\multicolumn{1}{c|}{\textbf{Methods}} & \multicolumn{1}{c|}{\textbf{RC@1}} & \multicolumn{1}{c|}{\textbf{RC@10}}  & \textbf{RC@50} & \textbf{AVG}
\\  \midrule[0.8pt]
  \midrule[0.8pt]
\textbf{FILM}~\cite{perez2018film}&10.19& 38.39& 68.30 &38.96  \\
\textbf{TRIG}~\cite{vo2019composing}&12.60& 45.45& 69.39& 42.48\\
\textbf{Relationship}~\cite{santoro2017simple}&12.31& 45.10& 71.45& 42.95 \\
\textbf{VAL}~\cite{chen2020image} &16.98& 49.83& 73.91& 46.91 \\
\textbf{ARTEMIS}~\cite{delmas2022artemis}& 17.60& 51.05& 76.85& 48.50\\
\textbf{SAC}~\cite{jandial2022sac} &18.11& 52.41& 75.42& 48.64\\
\textbf{ARTEMIS}$^{\dagger}$~\cite{delmas2022artemis} &\textcolor{red}{\textbf{18.72}}  &\textcolor{red}{\textbf{53.11}} &\textcolor{red}{\textbf{79.31}} &\textcolor{red}{\textbf{50.38}}\\ 
\midrule[0.8pt]
\textbf{ComposeAE}~\cite{anwaar2021compositional} & 4.37& 19.36& 47.58& 23.77\\
\textbf{SAC}$^{\dagger}$~\cite{jandial2022sac} &18.50 & 51.73 & \textcolor{blue}{\textbf{77.28}} & 49.17\\
\midrule[0.8pt]
\textbf{BOSS} (ours) &\textcolor{blue}{\textbf{18.70}}&\textcolor{blue}{\textbf{52.34}}&76.86&\textcolor{blue}{\textbf{49.30}}\\
\bottomrule[1.5pt]
\end{tabular}
\label{tab:results_3}
\end{table}

\begin{figure}[t]
\includegraphics[width=0.5\textwidth]{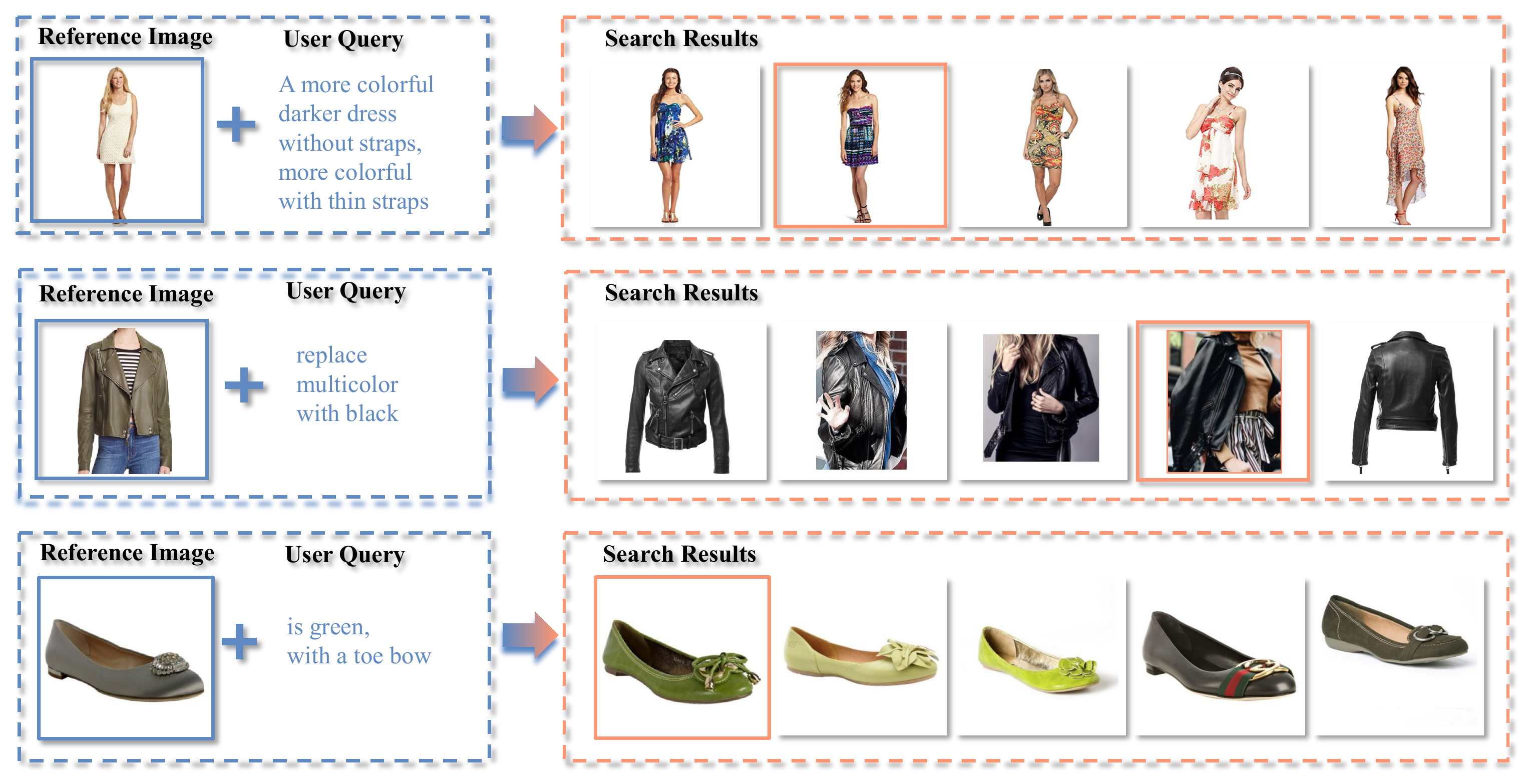}
\centering\caption{Qualitative examples from our BOSS on
FashionIQ, Fashion200k and Shoes datasets (top-down). \textcolor{blue}{Blue} box and \textcolor{myorange}{Orange} box refer to the composed query and  target image.}
\vspace{-0.2cm}
\label{fig:case}
\end{figure}

\noindent \textbf{Qualitative Results.} Figure~\ref{fig:case} presents the quantitative observations on three datasets. We observe that the BOSS is able to concurrently incorporate multiple semantic transformations in visual representations from text descriptions when retrieving images. On the one hand, 
the first case indicates that our method can search the \emph{target image} by learning the \emph{transformation}, \emph{i.e.},  changing certain attributes conditioned on \emph{query text} (\emph{e.g.}, ``white'' $\rightarrow$ ``more colorful darker'', ``wide strap'' $\rightarrow$  ``thin strap''). On the other hand, BOSS can capture the \emph{preservation} of the semantics in the \emph{reference image} and retrieve the corresponding target images based on these concepts (\emph{e.g.}, ``short dress'', ``has strap''  ).  As a summary, our approach can jointly
comprehend global appearance and local details within the composed query, aggregate the learned \emph{transformation} and \emph{preservation} to search the intended image correctly.

\subsection{In-Depth Analysis}
\label{sec:ida}
This section investigates the impact of different design choices in BOSS to further validate its effectiveness for CIR task.

\noindent \textbf{Importance of Individual Loss Component.}
We conduct an ablation study on FashionIQ dataset to illustrate the effect of each loss component in Table~\ref{tab:results_loss}, which indicates the following: Comparing BOSS and BOSS(-$\mathcal{L}_{hyb}$) (Row 2 vs Row 4), the hybrid counterfactual training significantly contributes 1.42\% and 4.40\% improvement on RC@10 and RC@50, respectively. Meanwhile, the results of Row 3 and Row 4 severally show the performance improvement of the reconstruct loss $\mathcal{L}_{res}$ and alignment loss $\mathcal{L}_{ali}$. Furthermore, the results of Row 1, which only adopt the unidirectional triplet loss from positive and negative target samples, obtained the worst performance.
These results both validate the importance and necessity of different loss modules for CIR accuracy improvement.

\begin{table}
\caption{\textbf{ Ablation study on FashionIQ dataset that showcase the impact of individual loss function.} }
\centering
\begin{tabular}[width=1\textwidth]{ccc|cc|c}
\toprule[1.5pt]  
\multicolumn{3}{c|}{\multirow{1}{*}{\textbf{Loss Functions}}}  &\multicolumn{2}{c|}{\multirow{1}{*}{\textbf{AVG}}}&\multicolumn{1}{c}{\multirow{2}{*}{\textbf{Total AVG}}}\\
\cmidrule(l){1-3}\cmidrule(l){4-5}
$\mathcal{L}_{hyb}$&  $\mathcal{L}_{res}$&$\mathcal{L}_{ali}$ &\multicolumn{1}{c}{\textbf{RC@10}}& \multicolumn{1}{c|}{\textbf{RC@50}}&
\\  \midrule[0.8pt]
  \midrule[0.8pt]
      & &  &26.46 &50.24&38.35  \\
    &\CheckmarkBold &\CheckmarkBold  &29.76 &55.46&42.61  \\
\CheckmarkBold & &\CheckmarkBold  &\textcolor{blue}{\textbf{30.29}} &\textcolor{blue}{\textbf{54.42}}&\textcolor{blue}{\textbf{42.36}} \\
\CheckmarkBold &\CheckmarkBold &  &28.18 &52.81&40.50\\
\CheckmarkBold &\CheckmarkBold &\CheckmarkBold  &\textcolor{red}{\textbf{31.61}} &\textcolor{red}{\textbf{ 56.06}}&\textcolor{red}{\textbf{ 43.85}}  \\

\bottomrule[1.5pt]
\end{tabular}
\label{tab:results_loss}
\end{table}

\noindent \textbf{Effect of Composition at Multi-level.} To build insights on the superiority of our bottom-up semantic composition module, we perform an in-depth analysis of search recall improvement on three datasets that benefits from the multi-level representation learning. Figure~\ref{fig:multi-level} summarizes the BOSS's performance using different level compositions. Evidently, taking coverage of compositional representation over both the local- and global- levels obtain the best retrieval performance. Either removing each of them will result in a performance decay, which proves the further effectiveness of our composition module. Moreover, this table also suggests that using global semantic composition can produce a comparable result, while taking utilization of simple local characteristic composition is insufficient to produce a satisfactory CIR performance. 

\begin{figure}[t]
\includegraphics[width=0.5\textwidth]{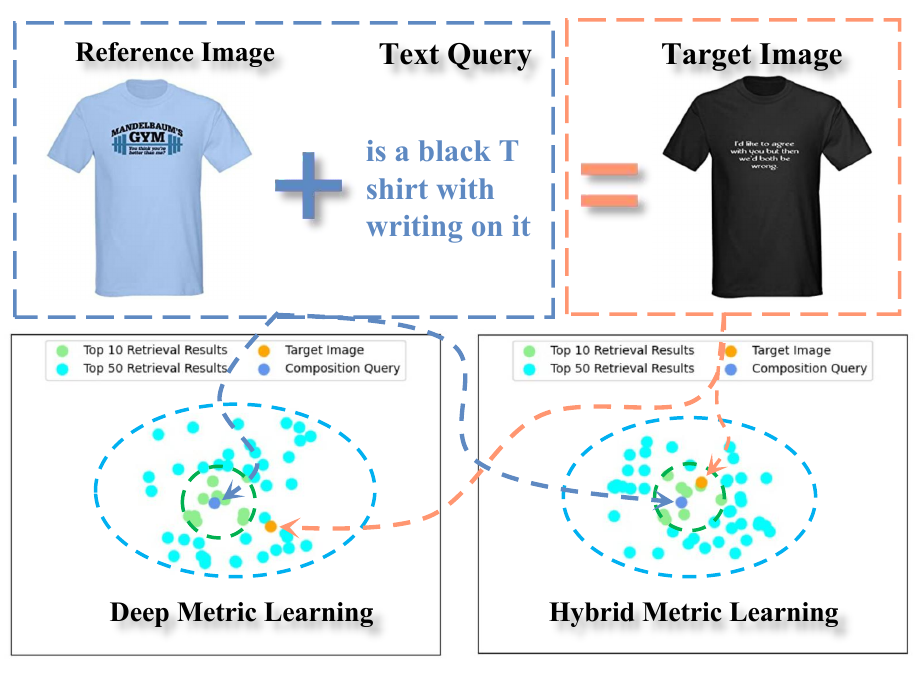}
\centering\caption{T-sne visualization shows the fine-grained image search that benefits from the hybrid counterfactual training. \textcolor{green}{Green} and \textcolor{blue}{blue} circles represent the RC@10 and RC@50 results in the image search semantic space, respectively.  }
\label{fig:vis}
\end{figure}

\begin{figure}[t]
\vspace{-0.4cm}
\includegraphics[width=0.5\textwidth]{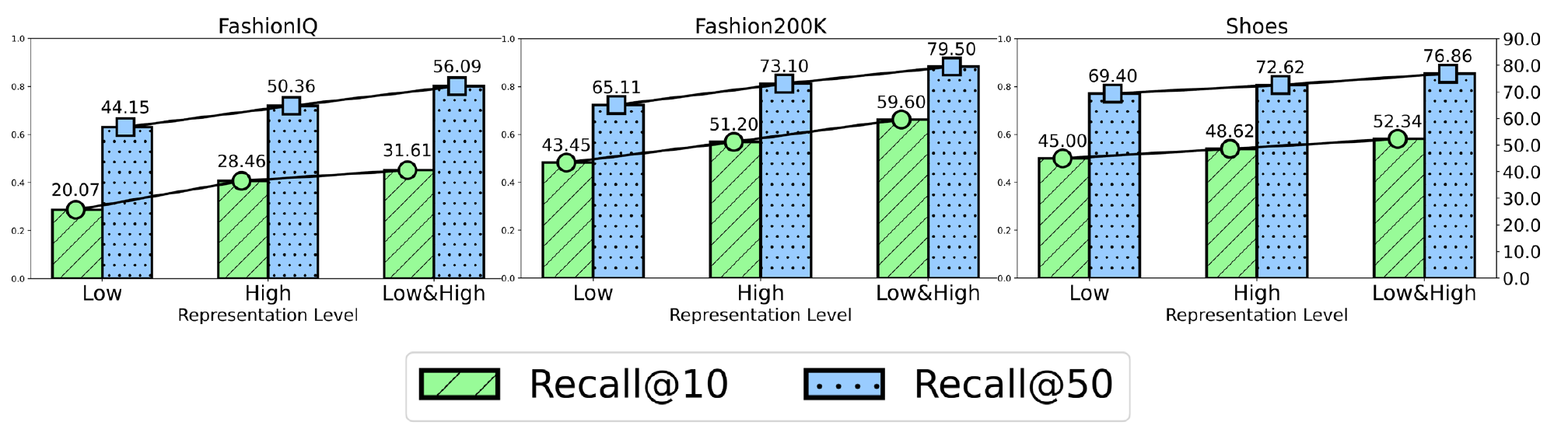}
\centering\caption{Effect of representation composition at multi-level.  }
\vspace{-0.3cm}
\label{fig:multi-level}
\end{figure}

\noindent \textbf{Benefits from Sensitivity Improvement.} 
We systematically present the explicit benefits from the hybrid counterfactual training(\emph{HCT}) and conduct an exploratory analysis. Figure~\ref{fig:vis} visualizes a qualitative example, while the deep metric learning (\emph{DML}) model searches the \emph{target image} at  RC@50 level, HML-based BOSS can retrieve it accurately at RC@10 level. We observe that the composed query (blue dot) and target representation (orange dot) is more closer in \emph{HCT} semantic space than \emph{DML}. After checking the dataset, we found that there are many composed queries has similar contexts but with different target images in FashionIQ. Under such circumstance, the \emph{DML} model may be confused which image is the optimal matching target. In contrast, the complete BOSS not only can guarantee the target representation is closer to the composite input query, but also force the target to pull away from the negative composition in the semantic space, \emph{i.e.}, similar queries. The bidirectionally optimized learning strategy is able to set up the fine-grained composition-target correspondence for accurate retrieval.

Broadly, most contemporary approaches  for
CIR prefer to learn a composed representation of the image-text
query pair and then measure the similarities of the global repre-
sentation with potential candidate images’ features to search for
the intended image. Despite intuitive, several critical factors for
CIR have been neglected by existing techniques: (1) Detailed local
characteristics utilization.

\section{Conlusion}
We introduced BOSS, a novel approach to tackle the challenging task of content-based image retrieval, which resolves the task into two major steps: an expressive visiolinguistic representation is derived from Bottom-up Semantic Composition, which electively preserves and transforms multi-level composition conditioned on text semantics; we develop a hybrid counterfactual training mechanism that is effective to construct the fine-grained composition-target correspondence by providing explicit supervised signals. We validate the efficacy of BOSS on three datasets. We hope our BOSS can provide a complement for existing literature of multi-modal retrieval task and benefit further study of vision and language.
\bibliographystyle{IEEEtran}
\bibliography{ref}

\vfill

\end{document}